\documentclass[10pt,twocolumn,letterpaper]{article}

\usepackage{iccv}
\usepackage{times}
\usepackage{epsfig}
\usepackage{graphicx}
\usepackage{amsmath}
\usepackage{amssymb}
\usepackage{multirow}

\usepackage[breaklinks=true,bookmarks=false]{hyperref}

\iccvfinalcopy % *** Uncomment this line for the final submission

\def\iccvPaperID{****} % *** Enter the ICCV Paper ID here

\ificcvfinal\fi

\begin{document}

%%%%%%%%% TITLE
\title{Motion Guided Attention for Video Salient Object Detection}

\author{\centerline{Haofeng Li$^{1}$ \hspace{1cm} Guanqi Chen$^{2}$ \hspace{1cm} Guanbin Li$^{2}\thanks{Corresponding author is Guanbin Li.}$ \hspace{1cm} Yizhou Yu$^{1,3}$}\\
\leftline{\hspace{0.7cm} $^1$The University of Hong Kong \hspace{0.7cm} $^2$Sun Yat-sen University \hspace{1cm} $^3$Deepwise AI Lab \hspace{1.5cm}}\\
{\tt\small lhaof@foxmail.com, chengq26@mail2.sysu.edu.cn, liguanbin@mail.sysu.edu.cn, yizhouy@acm.org}}

\maketitle
\ificcvfinal\thispagestyle{empty}\fi

%%%%%%%%% ABSTRACT
\begin{abstract}
   Video salient object detection aims at discovering the most visually distinctive objects in a video. How to effectively take object motion into consideration during video salient object detection is a critical issue. Existing state-of-the-art methods either do not explicitly model and harvest motion cues or ignore spatial contexts within optical flow images. In this paper, we develop a multi-task motion guided video salient object detection network, which learns to accomplish two sub-tasks using two sub-networks, one sub-network for salient object detection in still images and the other for motion saliency detection in optical flow images. We further introduce a series of novel motion guided attention modules, which utilize the motion saliency sub-network to attend and enhance the sub-network for still images. These two sub-networks learn to adapt to each other by end-to-end training. Experimental results demonstrate that the proposed method significantly outperforms existing state-of-the-art algorithms on a wide range of benchmarks. We hope our simple and effective approach will serve as a solid baseline and help ease future research in video salient object detection. Code and models will be made available.
\end{abstract}

%%%%%%%%% BODY TEXT
\section{Introduction}
Video salient object detection aims at discovering the most visually distinctive objects in a video, and identifying all pixels covering these salient objects. Video saliency detection tasks can be roughly categorized into two groups. The first group focuses on predicting eye fixations of viewers in a video, which may help biologically understand the inner mechanism of the human visual and cognitive systems. The second group requires the segmentation of the most important or visually prominent objects from a potentially cluttered background. In this paper, we attempt to address the second problem, namely, video salient object detection (SOD). A visual SOD model can serve as an important pre-processing component for many applications, for examples, image and video compression~\cite{itti2004automatic}, visual tracking~\cite{wu2014weighted} and person re-identification~\cite{zhao2013unsupervised}.

The most important difference between static images and videos is that objects in videos have motion, which is also a key factor that causes visual attention. That is, the motion of certain objects may make the object more prominent than others. How to effectively take object motion into consideration during video salient object detection is a critical issue for the following reasons.
First, object saliency in a video is not only determined by object appearance (including color, texture and semantics), but also affected by object motion between consecutive frames. Itti \textit{et al.} \cite{itti1998a} suggest that differences between consecutive frames resulting from object motion are more attractive to human attention.
Second, object motion provides an essential hint on spatial coherence. Neighboring image patches with similar displacements very possibly belong to the same foreground object, or the background region.
Third, exploiting motion cues makes the segmentation of salient objects in a video easier, and hence, produces saliency maps of higher quality. For example, in RGB frames, the background may contain diverse contents with different colors and texture, and the foreground object may be composed of parts with sharp edges and different appearances. It is challenging to locate and segment complete salient objects in such video frames without motion cues.
\begin{figure}[!t]
\begin{center}
\includegraphics[width=1.00\linewidth]{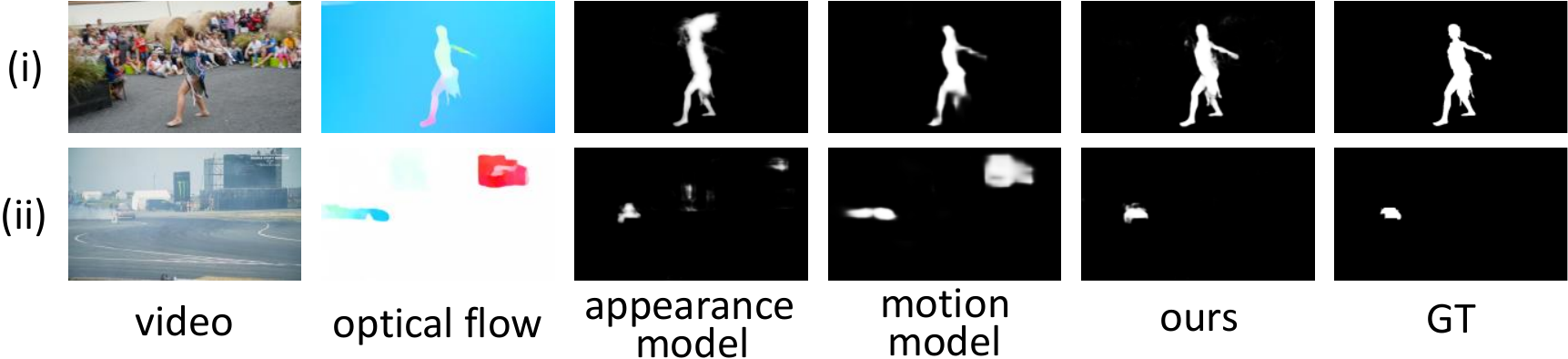}
\end{center}
\caption{Effectiveness of our proposed network. An appearance based saliency model and a motion based saliency model, which take an RGB video frame and an optical flow image as their input respectively, have their own advantages and weaknesses. Our proposed method successfully complements an appearance branch with a motion branch, and outperforms any one of them.}
\label{fig:firstSample}
\end{figure}

Video saliency detection have attracted a wide range of research interests in the field of computer vision. However, existing video SOD algorithms have not sufficiently exploited the properties of object motion. Graph based methods~\cite{wang2015consistent,chen2017video,liu2017saliency} intend to combine appearance saliency with motion cues on the basis of spatio-temporal coherence, but they are limited by the use of handcrafted low-level features and the lack of training data mining. Thus such graph based algorithms fail to adaptively harvest accurate features for motion patterns and object semantics in complicated scenes. It is arduous for these methods to capture the contrast and uniqueness of object motion and high-level semantics. Fully convolutional network based methods~\cite{sun2018sg,wang2018video} model temporal coherence by simply concatenating past frames or past predicted saliency maps with the current frame to form the input of convolutional neural networks (CNN). These CNN based methods do not employ explicit motion estimation, such as optical flow, and are affected by the distractive and cluttered background from the video appearance. Currently, state-of-the-art results of video salient object detection are achieved by recurrent neural network based algorithms~\cite{li2018flow,song2018pyramid}, which exploit convolutional memory units such as ConvLSTM to aggregate long-range spatio-temporal features. Some of these recurrent models~\cite{li2018flow} make use of flow warping to align previous features with the current one, but overlook the spatial coherence and motion contrast within an optical flow image.

Motivated by the above observations, in this paper, we propose a multi-task motion guided video salient object detection network, which models and exploits motion saliency to identify the salient objects in a video. To explicitly investigate how motion contrast influences video saliency, we partition the video salient object detection task into two sub-tasks, salient object detection in a static image, and motion saliency inferred from an optical flow image. We first carry out these two sub-tasks with two separate branches. Then we integrate these two branches together to accomplish the overall task. Specifically, the proposed method attends the branch for static images with motion saliency produced from the branch for optical flow images to compute the overall saliency of video objects. Moreover, to implement the above attention mechanism, we develop a set of novel motion guided attention modules, which aggregate the advantages of residual learning as well as spatial and channel-wise attention.

We claim that the proposed method is a strong baseline, which does not need long-range historical features as ConvLSTM based algorithms~\cite{le2018video,wang2018video}, but only requires short-range contexts computed from the previous frame. In short, the contributions of this paper are summarized as follows.
\begin{itemize}
\item We introduce a collection of novel motion guided attention modules, which can attend and enhance appearance features with motion features or motion saliency.
\item We develop a novel network architecture for video salient object detection. The proposed network is composed of an appearance branch for salient object detection in still images, a motion branch for motion saliency detection in optical flow images, and our proposed attention modules bridging these two branches.
\item Extensive experiments are conducted to verify the effectiveness of the proposed attention modules and the proposed network. Experimental results indicate that our proposed method significantly surpasses existing state-of-the-art algorithms on a wide range of datasets and metrics.
\end{itemize}

\section{Related Work}

\subsection{Video Salient Object Detection}
Many video salient object detection methods~\cite{wang2015consistent,wang2015saliency,liu2017saliency,wang2018video,li2018flow,song2018pyramid,le2018video,Fan_2019_CVPR} have been studied recently. In particular, deep learning based video SOD algorithms have achieved significant success, and fall into two categories, region-wise labeling, and pixel-wise labeling. STCRF~\cite{le2018video} extracts deep features for image regions, and proposes a spatiotemporal conditional random field to compute a saliency map based on region-wise features. Dense labeling models for video SOD are also divided into two main types, one using fully convolutional network (FCN), and the other embracing recurrent neural network. FCNS~\cite{wang2018video} employs a static saliency FCN that predicts a saliency map based on current frame, and a dynamic saliency FCN which takes the predicted static saliency, the current and the next frame as input to produce the final result. FGRNE~\cite{li2018flow} utilizes a ConvLSTM to refine former optical flows, warps former visual features with refined flows, and adopts another ConvLSTM to aggregate former and current features. PDB~\cite{song2018pyramid} employs two parallel dilated  bi-directional ConvLSTMs to implicitly discover long-range spatio-temporal correlations, but disregards explicit distinctive motions and how they affect object saliency in a video.

\subsection{Visual Attention Model}
Attention mechanisms, which highlight different positions or nodes according to their importance, have been widely adopted in the field of computer vision. Xu \textit{et al.} develop an image caption model~\cite{xu2015show} based on stochastic hard attention and deterministic soft attention. Wang \textit{et al.} propose a residual attention network~\cite{wang2017residual} built on stacked residual attention modules, to solve image classification tasks. Fu \textit{et al.} introduce a recurrent attention convolutional neural network (RA-CNN)~\cite{fu2017look} which recursively explores discriminative spatial regions and harvests multi-scale region based features for fine-grained image recognition. Wu \textit{et al.} propose to employ a structured attention mechanism to integrate local spatial-temporal representation at trajectory level~\cite{wu2018interpretable} for more fine-grained video description.
In this paper, we are the first to explore the complementary enhancement effect of motion information on appearance contrast modeling from the perspective of various  attention schemes.

\subsection{Motion based Modeling}
Optical flow represents pixel-level motion between two consecutive frames in a video. The following briefs some popular optical flow estimation methods~\cite{dosovitskiy2015flownet,ilg2017flownet,sun2018pwc-net}, and their applications in motion based modeling~\cite{jain2017fusionseg,tokmakov2017learningmotion,tokmakov2017learning}.
Dosovitskiy \textit{et al.}~\cite{dosovitskiy2015flownet} calculate optical flows by concatenating two consecutive frames as input and harvesting patch-wise similarities between two frames.
FlowNet 2.0~\cite{ilg2017flownet} employs two parallel streams to estimate small and large displacements respectively, and fuses them at last.
Fusionseg~\cite{jain2017fusionseg} adopts an appearance stream and a motion stream to model video segmentation, but simply fuses them with element-wise multiplication and maximum. Tokmakov \textit{et al.}~\cite{tokmakov2017learning} also utilize a dual-stream architecture and attempt to fuse two streams via concatenation and a convolutional memory unit (ConvGRU). Existing motion based deep learning methods lack investigating how motion cues (particularly, motion saliency) affect appearance features as well as object saliency in an attention manner.

\section{Method}

\subsection{Motion Guided Attention}\label{sec:MGA}
Let us consider how to exploit motion information to emphasize some important positions or elements in an appearance feature. We define an \textit{appearance feature} as a feature tensor generated by some hidden layers such as some ReLU functions in the appearance branch. The motion information can be categorized into two groups. The first group denotes motion saliency maps that are yielded by the last layer in the motion branch. Such motion saliency maps can be predicted with a Sigmoid activation function and hence their elements are within the range of $[0,1]$. The second group represents motion features that are produced by some intermediate ReLU functions inside the motion sub-network.

\begin{figure}[t]
\begin{center}
\includegraphics[width=0.9\linewidth]{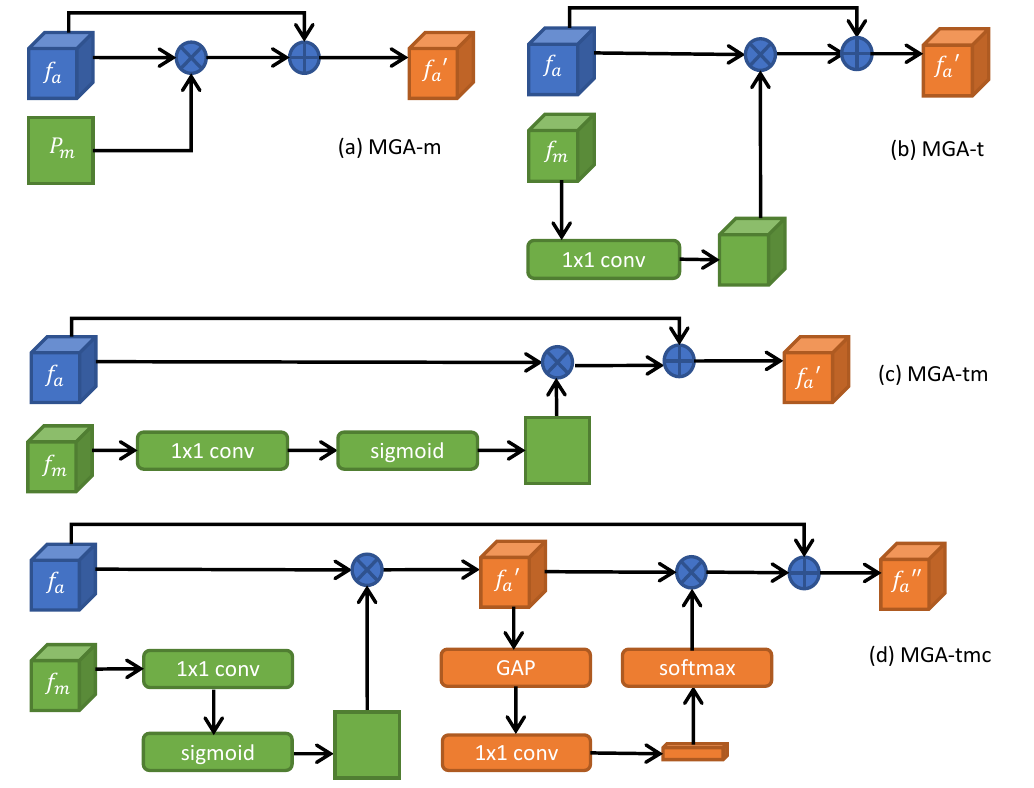}
\end{center}
\caption{Motion Guided Attention Modules}
\label{fig:MotionGuidedAttention}
\end{figure}

Consider a simple case, utilizing a motion saliency map to attend an appearance feature. The motion saliency map is denoted as $P_m$ (the Prediction of the Motion branch) and the appearance feature is denoted as $f_a$. A straightforward way for computing attended appearance feature $f'_a$ is
$f'_a = f_a \otimes P_m$,
where $f'_a$, $f_a$ and $P_m$ are of size $C\times H\times W$, $C\times H\times W$ and $H\times W$ respectively. $\otimes$ denotes element-wise multiplication, namely, applying element-wise multiplication between $P_m$ and each channel slice of $f_a$. Such multiplication based attention is simple but has limitations. Since the motion branch is trained with a motion saliency detection task, image parts which has similar displacement with the background are most likely predicted as 0 in $P_m$. Consider that only some parts of a salient object move in some video frame, as shown in Figure~\ref{fig:firstSample}(i). Then the still parts of the salient object could be 0 in $P_m$ and hence their corresponding features in $f'_a$ are suppressed. In such case the naive multiplication attention fails to maintain the complete salient object. To alleviate the above issue, we propose a variant that is not to `block out' unsalient-motion regions but only to highlight salient-motion regions, formulated as:
\begin{equation}\label{eq:MGA_m}
f'_a = f_a \otimes P_m + f_a
\end{equation}
where $+$ denotes element-wise additions. The multiplication based attention serves as a residual term in Eq~(\ref{eq:MGA_m}). The additional term $+ f_a$ complements the features that may be incorrectly suppressed by $f_a\otimes P_m$. Thus the residual formulation is promising to attend salient-motion parts without discarding still but salient areas. We name the proposed attention module in Eq~(\ref{eq:MGA_m}) and Figure~\ref{fig:MotionGuidedAttention}(a) as MGA-m. MGA denotes Motion Guided Attention and `-m' means that the motion input of the attention module is a map.

The following discusses how to employ a motion feature tensor $f_m$ to draw attentions to some elements in an appearance feature $f_a$. Consistent with MGA-m using a multiplication-and-addition manner, we first propose a motion guided attention module with two tensor inputs in the below:
\begin{equation}\label{eq:MGA_t}
    f'_a = f_a \otimes g( f_m ) + f_a
\end{equation}
where $f_a$ and $f_m$ are of size $C\times H\times W$ and $C'\times H\times W$ respectively. $g(\cdot)$ is a $1\times 1$ convolution which aligns the shape of the motion feature with that of the appearance feature. Then an attention mechanism in an element-wise multiplication-and-addition way is applicable, between the appearance feature and the output of $g(\cdot)$. The proposed motion guided attention module shown in Eq~(\ref{eq:MGA_t}) and Figure~\ref{fig:MotionGuidedAttention}(b) is dubbed as MGA-t in which `-t' means that the input from motion branch is a feature tensor.

Inspired by the MGA-m module that exploits the motion information as spatial attention weights, we conceive a variant to attend a tensor with the other one, by converting the motion feature into spatial weights beforehand. Such attention module can be formulated as :
\begin{equation}\label{eq:MGA_tm}
    f'_a = f_a \otimes \mbox{Sigmoid}( h( f_m ) ) + f_a
\end{equation}
where $h(\cdot)$ denotes a $1\times 1$ convolution with 1 output channel. Thus the output of $\mbox{Sigmoid}(\cdot)$ is an attention map of size $H\times W$. The above module shown in Eq~(\ref{eq:MGA_tm}) and Figure~\ref{fig:MotionGuidedAttention}(c) is named MGA-tm in which `-tm' means that the input feature tensor from motion branch is transformed to a spatial map at the very beginning. Let us discuss the difference between the MGA-t module and the MGA-tm module. The MGA-tm module can be viewed as applying spatial attention with the motion feature, while in the MGA-t module spatial and channel-wise attention is implemented at the same time via a 3D tensor of attention weights. Note that in our proposed method, the motion branch only takes an optical flow image as input, serves as passing messages towards the appearance branch, and has no knowledge of appearance information. Thus it may be not so promising to achieve channel-wise attention with the motion feature alone. However, for the MGA-tm module, it lacks emphasizing important channels that is closely associated with visual saliency or salient-motion objects. Based on these considerations, we come up with the fourth MGA module as:
\begin{align}\label{eq:MGA_tmc_spatial}
    f'_a &= f_a \otimes \mbox{Sigmoid}(h(f_m)),\\
    f''_a &= f'_a \otimes [ \mbox{Softmax}( h'( \mbox{GAP}( f'_a ) ) ) \cdot C ] + f_a \label{eq:MGA_tmc_channel}
\end{align}
where $f_a$, $f'_a$ and $f''_a$ all are tensors of size $C\times H\times W$. $f_m$ is a $C'\times H\times W$ tensor. Both $h(\cdot)$ and $h'(\cdot)$ are implemented as $1\times 1$ convolutions whose output channels are 1 and $C$ respectively. $\mbox{GAP}(\cdot)$ denotes global average pooling in the spatial dimensions. $C$ in Eq~(\ref{eq:MGA_tmc_channel}) is a single scalar and equals to the number of elements in the output of the Softmax function. The proposed motion guided attention module shown in Eq~(\ref{eq:MGA_tmc_spatial}-\ref{eq:MGA_tmc_channel}) and Figure~\ref{fig:MotionGuidedAttention}(d) is named MGA-tmc where the last `c' represents channel-wise attention.
\begin{figure*}[!t]
\begin{center}
\includegraphics[width=0.9\linewidth]{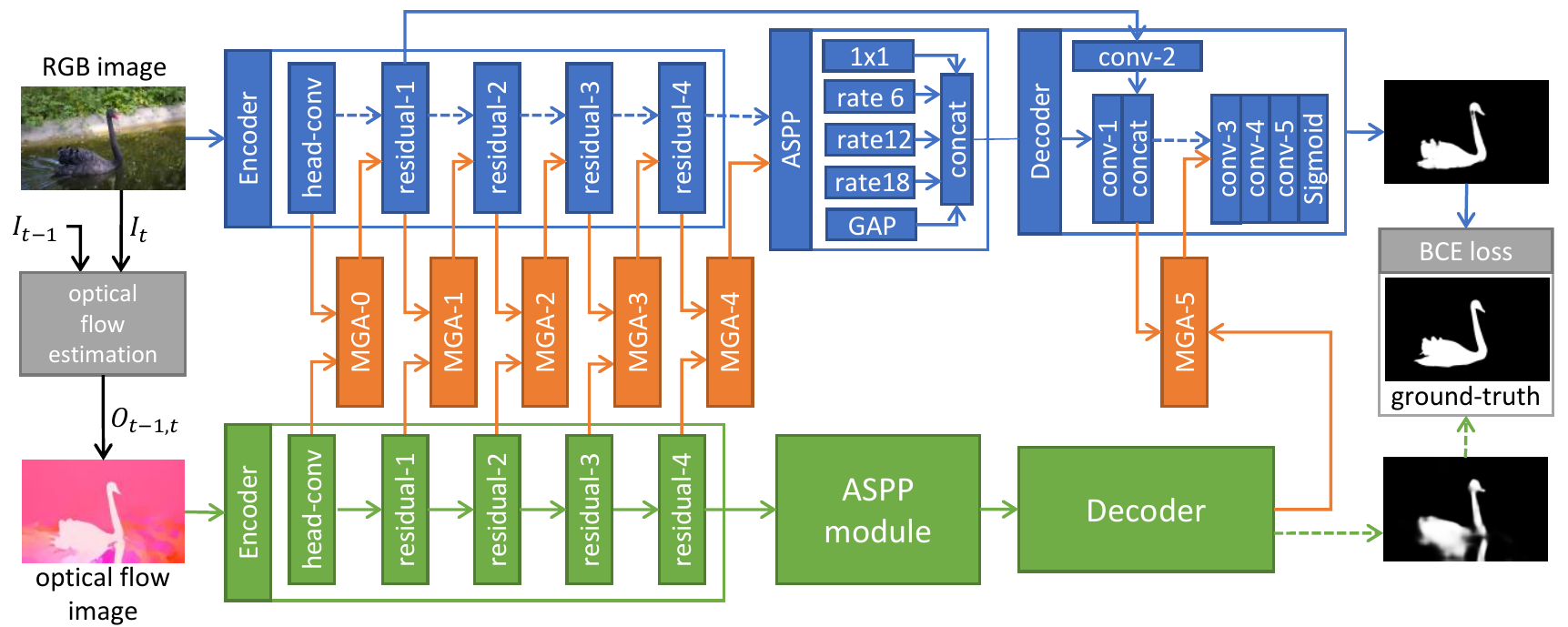}
\end{center}
\caption{Motion Guided Attention Network. The blue parts denote the appearance branch while the green parts represent the motion branch. The blue dashed lines are connected when the appearance branch solves the static-image SOD sub-task alone. The green dashed lines are linked when the motion branch alone is trained on the motion-image SOD sub-task. The orange parts are the proposed motion guided attention modules, and work only while addressing the video SOD task. BCE loss denotes binary cross-entropy loss.}
\label{fig:MultiTaskNetwork}
\end{figure*}

Let us present more rationales behind the MGA-tmc module. $f'_a$ is an appearance feature already spatially highlighted by a motion feature. $\mbox{GAP}(f'_a)$ harvests a global representation of $f'_a$ and outputs a single vector of $C$ elements. Based on the global representation, $h'(\cdot)$ predicts a vector of $C$ scalar weights for channels. These channel-wise attention weights aim at selecting or strengthening the responses of essential attributes such as some kind of edges, boundaries, colors, texture and semantics. $\mbox{Softmax}(\cdot)\cdot C$ normalizes the output of $h'(\cdot)$ such that the mean value of the attention weights equals to 1. For simplicity, $\mbox{Softmax}(\cdot)\cdot C$ is denoted as `softmax' in Figure~\ref{fig:MotionGuidedAttention}(d), and multiplying by $C$ is omitted. $f'_a \otimes [\cdot]$ in Eq~(\ref{eq:MGA_tmc_channel}) is to multiply the feature column at each spatial position of $f'_a$ by the normalized attention vector. To summarize, the MGA-tmc module first emphasizes the spatial locations with salient motions, then selects attributes which is potential to model saliency conditioned on the motion-attended appearance features, and finally adds the input feature as a complement. The effectiveness of our proposed attention modules (MGA-m, MGA-t, MGA-tm and MGA-tmc) will be validated in Section~\ref{sec:exp}.

\subsection{Network Architecture}
As shown in Figure~\ref{fig:MultiTaskNetwork}, our proposed network architecture consists of an appearance branch, a motion branch, a pretrained flow-estimation network and a set of motion guided attention modules bridging the appearance and the motion branch. The flow-estimation network denoted as `optical flow estimation' in Figure~\ref{fig:MultiTaskNetwork} is implemented as \cite{ilg2017flownet}. The architectures of the appearance sub-network and the motion sub-network are quite similar but different. 
The motion sub-network utilizes a lighter design than the appearance one, since the optical flow image does not contain as much high-level semantics and subtle boundaries as the RGB image.

The proposed method divides a video salient object detection task into two sub-tasks, appearance based static-image saliency detection and motion saliency detection. We first introduce the architectures of the appearance sub-network and the motion sub-network during separate training. Both the appearance branch and the motion branch are composed of three parts, an encoder, an atrous spatial pyramid pooling (ASPP) module and a decoder. The encoder works by extracting low-level to high-level visual features and reducing the resolution of feature maps. The encoder includes five layers: a head-convolution and four residual layers denoted as residual-\textit{i} ($i \in \{1,2,3,4\}$). The head-convolution has 64 output channels, $7\times 7$ kernel size and a stride of 2, followed by a batch normalization and a ReLU function. For the appearance branch, these four residual layers contain 3, 4, 23 and 3 residual learning based `bottlenecks'~\cite{he2016deep}, and have 256, 512, 1024 and 2048 output channels respectively. For the motion branch, its residual layers adopt 3, 4, 6 and 3 basic residual learning blocks~\cite{he2016deep}, and have 64, 128, 256 and 512 output channels respectively. The strides of these four residual layers are set as 2, 2, 1 and 1 respectively in both sub-networks. Thus the encoder reduces the spatial size of input feature map as 1/8 of the original size.

The ASPP module harvests long-range dependencies within a feature map via dilated convolutions, and integrates them with local and global representations, which could implicitly capture long-range contrast for saliency modeling. As shown in Figure~\ref{fig:MultiTaskNetwork}, the ASPP module passes the input feature through five parallel layers which are a $1\times 1$ pointwise convolution, three $3\times 3$ convolutions with dilation rate of 12, 24 and 36 respectively, and a global average pooling layer. The outputs of these five parallel layers are concatenated along with the dimension of depth, which yields a single feature map.

The decoder recovers the spatial size of feature maps to predict high-resolution saliency maps with accurate object boundaries, by fusing the low-level feature and the high-level one together. As Figure~\ref{fig:MultiTaskNetwork} displays, the high-level output of the ASPP module is shrinked to a 256-channels feature via a $1\times 1$ convolution `conv-1' in the decoder, while the low-level output from residual-1 is reduced to a 48-channels feature by another $1\times 1$ convolution `conv-2'. After concatenating the low-level feature with the high-level one, two $3\times 3$ convolutions denoted as `conv-3' and `conv-4' with 256 output channels follows. Next, a $1\times 1$ convolution `conv-5' followed by a Sigmoid function predicts the final single-channel saliency map. For simplicity, the decoder of motion branch uses three layers similar to conv-\{3-5\} to directly infer a motion saliency map.

Importantly, let us introduce how to adapt the appearance branch and the motion branch to our proposed motion guided attention modules for video salient object detection. As can be seen in Figure~\ref{fig:MultiTaskNetwork}, MGA-$i$ ($i \in \{0, 1, 2, 3, 4, 5\}$) represents six attention modules in our proposed multi-task network. MGA-0 takes the outputs of two head-convolutions from the appearance sub-network and the motion sub-network, as its inputs. MGA-$i$ takes the output features of residual-$i$ from the two branches, as its inputs. Note that in the appearance sub-network, the direct linkages among five layers within its encoder are removed. The output of MGA-0 replaces that of the head-conv to be passed into the residual-1 layer in the appearance branch. Similarly, residual-\textit{i} in the appearance branch uses the output produced by MGA-($i$-1) instead of residual-($i$-1), as its input. MGA-4 takes the place of residual-4 to be connected with the ASPP module in the appearance sub-network. Different from the appearance branch, the encoder in the motion branch still maintains its internal linkages and provides side-outputs as the input of MGA-\{0-4\}. MGA-\{0-4\} are located at the encoder side while MGA-5 works at the decoder side. MGA-5 employs the final output of the motion branch, and the fusion of the low and high-level features in the appearance branch, as its inputs. The output of MGA-5 also replaces the fused feature to be passed into `conv-3' in the appearance sub-network. Since the motion input of MGA-5 is a single-channel saliency map, it only can be instantiated with the MGA-m module. As for MGA-\{0-4\}, their implementations could be selected among MGA-t, MGA-tm and MGA-tmc.
\begin{table*}[!t]
%\small
%\footnotesize
%\scriptsize
\tiny
\begin{center}
\resizebox{0.9\linewidth}{!}{
    \begin{tabular}{|l|r|rrr|rrr|rrr|}
    \hline
    \multirow{2}{*}{Methods}& \multirow{2}{*}{Year}&         & DAVIS &       &       & FBMS  &       &       & ViSal &       \\
                            &                      & MAE     & S-m    & maxF  & MAE   & S-m    & maxF  & MAE   & S-m    & maxF  \\
    \hline
    Amulet~\cite{zhang2017amulet} & ICCV'17  & 0.109   & 0.748 & 0.719 & 0.133 & 0.753 & 0.746 & 0.058 & 0.874 & 0.888 \\
    UCF~\cite{zhang2017learning}  & ICCV'17  & 0.164   & 0.698 & 0.742 & 0.195 & 0.708 & 0.718 & 0.119 & 0.798 & 0.880 \\
    SRM~\cite{wang2017stagewise}  & ICCV'17  & {\color{blue}0.040}   & 0.840 & 0.795 & 0.073 & 0.805 & 0.792 & 0.028 & 0.914 & 0.916 \\
    DSS~\cite{hou2017deeply}      & CVPR'17  & 0.047   & 0.827 & 0.773 & 0.081 & 0.799 & 0.785 & 0.026 & 0.927 & 0.921 \\
    MSR~\cite{li2017instance}     & CVPR'17  & 0.062   & 0.798 & 0.762 & 0.081 & 0.810 & 0.792 & 0.045 & 0.892 & 0.890 \\
    NLDF~\cite{luo2017non}        & CVPR'17  & 0.059   & 0.803 & 0.760 & 0.085 & 0.794 & 0.771 & 0.022 & 0.925 & 0.920 \\
    R3Net~\cite{deng2018r3net}    & IJCAI'18 & 0.064   & 0.786 & 0.746 & 0.090 & 0.790 & 0.759 & 0.025 & 0.921 & 0.911 \\
    C2SNet~\cite{li2018contour}   & ECCV'18  & 0.052   & 0.813 & 0.771 & 0.073 & 0.811 & 0.782 & 0.023 & 0.922 & 0.924 \\
    RAS~\cite{chen2018reverse}    & ECCV'18  & 0.057   & 0.785 & 0.729 & 0.078 & 0.816 & 0.807 & {\color{green}0.019} & {\color{blue}0.930} & 0.925 \\
    %RSDNet   & CVPR'18  & 0.134   & 0.616 & 0.466 & 0.096 & 0.769 & 0.725 & 0.050 & 0.878 & 0.866 \\
    DGRL~\cite{wang2018detect}    & CVPR'18  & 0.056   & 0.812 & 0.763 & {\color{green}0.057} & 0.829 & 0.802 & 0.022 & 0.916 & 0.917 \\
    PiCANet~\cite{liu2018picanet} & CVPR'18  & 0.044   & {\color{blue}0.842} & {\color{blue}0.801} & {\color{blue}0.059} & {\color{blue}0.845} & {\color{blue}0.819} & 0.022 & {\color{green}0.937} & {\color{blue}0.932} \\
    \hline
    GAFL~\cite{wang2015consistent}& TIP'15   & 0.122   & 0.697 & 0.658 & 0.199 & 0.615 & 0.575 & 0.101 & 0.774 & 0.759 \\
    SAGE~\cite{wang2015saliency}  & CVPR'15  & 0.137   & 0.648 & 0.569 & 0.192 & 0.624 & 0.598 & 0.094 & 0.781 & 0.771 \\
    SGSP~\cite{liu2017saliency}   & TCSVT'17 & 0.143   & 0.678 & 0.707 & 0.211 & 0.590 & 0.601 & 0.171 & 0.694 & 0.682 \\
    FCNS~\cite{wang2018video}     & TIP'18   & 0.056   & 0.802 & 0.750 & 0.103 & 0.775 & 0.763 & 0.041 & 0.897 & 0.892 \\
    FGRNE~\cite{li2018flow}       & CVPR'18  & 0.044   & 0.838 & 0.797 & 0.078 & 0.814 & 0.794 & 0.049 & 0.871 & 0.845 \\
    PDB~\cite{song2018pyramid}    & ECCV'18  & {\color{green}0.029} & {\color{green}0.879} & {\color{green}0.862} & 0.070 & {\color{green}0.846} & {\color{green}0.829} & {\color{blue}0.021} & 0.928 & {\color{green}0.936} \\
    ours     &          & {\color{red}0.022}   & {\color{red}0.913} & {\color{red}0.902} & {\color{red}0.027} & {\color{red}0.907} & {\color{red}0.910} & {\color{red}0.015} & {\color{red}0.944} & {\color{red}0.947} \\
    \hline
    \end{tabular}
}
\end{center}
\caption{Comparisons with state-of-the-art video salient object detection algorithms. The three best performing algorithms are marked in {\color{red}red}, {\color{green}green}, and {\color{blue}blue} respectively. }
\label{tab:videoSODComp}
\end{table*}

\subsection{Multi-task Training Scheme}
\begin{figure*}[!h]
	\begin{center}
		\includegraphics[width=0.9\linewidth]{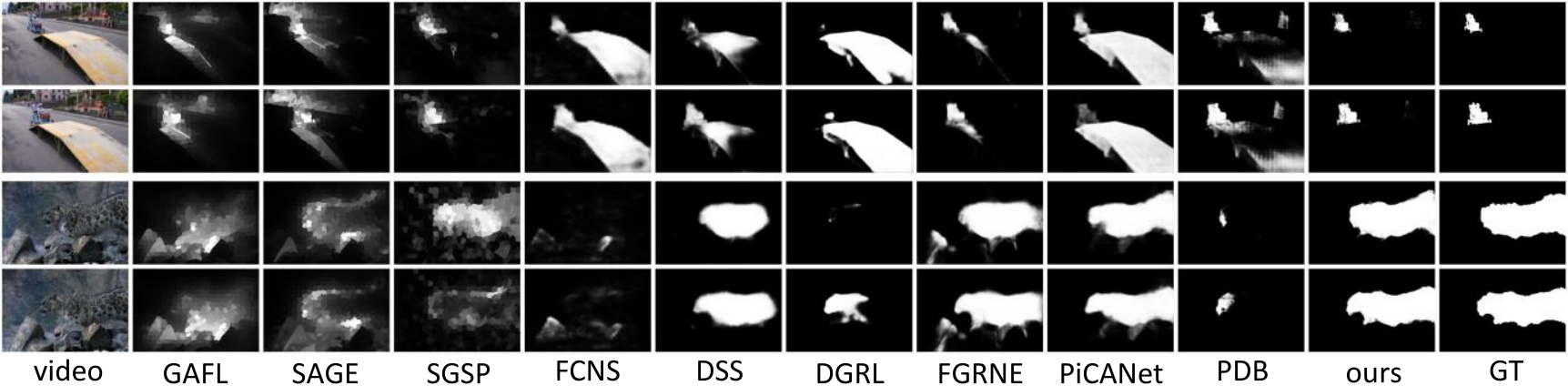}
		%\leftline{\hspace{1.1cm}video \hspace{0.4cm}GAFL \hspace{0.3cm}SAGE \hspace{0.35cm}SGSP \hspace{0.4cm}FCNS \hspace{0.4cm}DSS \hspace{0.5cm}DGRL \hspace{0.15cm}FGRNE \hspace{0.05cm}PiCANet \hspace{0.25cm}PDB \hspace{0.5cm}ours \hspace{0.7cm}GT}
	\end{center}
	\caption{Qualitative comparison with state-of-the-art video salient object detection methods.}
	\label{fig:QualityComp}
\end{figure*}
We develop a multi-task pipeline to train our proposed motion guided attention network. First, we initialize the appearance sub-network using a ResNet-101~\cite{he2016deep} pretrained on ImageNet~\cite{deng2009imagenet,russakovsky2015imagenet}, and then fine-tune the appearance branch on a static-image salient object detection dataset. Second, we implement the `optical flow estimation'~\cite{ilg2017flownet}, and employ it to render optical flow images according to~\cite{Butler:ECCV:2012} on our training set of video salient object detection. The optical flow images are computed as a forward flow from the previous frame to the current frame. Third, the motion sub-network is initialized using an ImageNet-pretrained ResNet-34~\cite{he2016deep} model, and then is trained on these synthesized optical flow images and their corresponding saliency maps in the video salient object detection dataset. Lastly, the proposed MGA modules integrate  the two branches to form our proposed network, which is tuned with a mixture of static-image and video salient object detection datasets. Since training samples of a static image or the first frame in an video have no corresponding motion images, we assume that their previous frame are as the same as themselves. That is to say, objects in these samples are not in motion and no salient motions exist. For such cases, we simply fill zeros in the motion inputs of the MGA modules.

\section{Experiments}\label{sec:exp}
In this paper, we choose the train set of DUTS~\cite{wang2017learning}, DAVIS~\cite{perazzi2016benchmark} and FBMS~\cite{brox2010object} as our training set. We evaluate video salient object detection methods on DAVIS, FBMS and ViSal~\cite{wang2015consistent} benchmark. DUTS is a commonly used static-image salient object detection dataset. ViSal dataset can be used to rate the generalization of video salient object detection models, since all video SOD algorithms are not trained with any subsets of ViSal. Mean absolute error (MAE), structure-measure (S-m)~\cite{fan2017structure}, max F-measure (maxF)~\cite{achanta2009frequency}, Precision-Recall (P-R) curves and Fmeasure-Threshold curves are selected as criteria. The results of PR curves and Fmeasure-Threshold curves can be found in the supplemental materials. SGD algorithm is used to train the proposed network with an initial learning rate of 10$^{-8}$, a weight decay of 0.0005 and an momentum of 0.9. The proposed method costs about 0.07 seconds for a single frame, regardless of flow estimation.

\subsection{Comparison with the state-of-the-art}
As shown in Table~\ref{tab:videoSODComp}, our proposed method is compared with 11 existing static-image salient object detection models including Amulet~\cite{zhang2017amulet}, UCF~\cite{zhang2017learning}, SRM~\cite{wang2017stagewise}, DSS~\cite{hou2017deeply}, MSR~\cite{li2017instance}, NLDF~\cite{luo2017non}, R3Net~\cite{deng2018r3net}, C2SNet~\cite{li2018contour}, RAS~\cite{chen2018reverse}, DGRL~\cite{wang2018detect}, PiCANet~\cite{liu2018picanet}, and 6 state-of-the-art video SOD algorithms including GAFL~\cite{wang2015consistent}, SAGE~\cite{wang2015saliency}, SGSP~\cite{liu2017saliency}, FCNS~\cite{wang2018video}, FGRNE~\cite{li2018flow}, PDB~\cite{song2018pyramid}. Our proposed method is implemented by adopting MGA-tmc module at the positions of MGA-\{0-4\}, and MGA-m module at the position of MGA-5. The results of our method in Table~\ref{tab:videoSODComp} are obtained without any post-processing. We utilize the public released code and pretrained weights of PDB whose performance is slightly higher than its original paper~\cite{song2018pyramid}. As Table~\ref{tab:videoSODComp} displays, the proposed method achieves the lowest MAE, the highest S-m and maxF on all three benchmarks DAVIS, FBMS and ViSal. On the DAVIS dataset, the proposed method considerably outperforms the second best model PDB by 3.4\% S-m and 4.0\% maxF. On the FBMS benchmark, our algorithm significantly surpasses the second best method PDB by 6.1\% S-m and 8.1\% maxF. The proposed network also obtains 3.0\% MAE smaller than the second best algorithm DGRL on FBMS. As for the ViSal dataset, our proposed method demonstrates 0.7\% S-m and 1.1\% maxF higher than the second best models PiCANet and PDB respectively. Since ViSal is a relatively small and easy benchmark in comparison to DAVIS and FBMS, the numeric results of state-of-the-art methods including ours are close. ViSal does reflect the generalization capacity of video SOD models for none of existing methods is trained with videos from the ViSal dataset. Thus, our proposed method not only establishes a new state-of-the-art for the video salient object detection task, but is also promising to enjoy superior generalization in real applications. Figure~\ref{fig:QualityComp} presents a qualitative comparison between the state-of-the-art algorithms and the proposed network. More qualitative results are placed in the supplemental materials.

As displayed in Table~\ref{tab:segmentationComp}, the proposed method is compared with 9 latest unsupervised video segmentation algorithms including SAGE~\cite{wang2015saliency}, LVO~\cite{tokmakov2017learning}, FSEG~\cite{jain2017fusionseg}, ARP~\cite{koh2017primary}, PDB~\cite{song2018pyramid}, MSGSTP~\cite{hu2018unsupervised}, MBN~\cite{li2018unsupervised}, IET~\cite{li2018instance} and MotAdapt~\cite{siam2018video}. To assess the performance of these models, we resort to widely used evaluation metrics, $\mathcal{J}$ Mean, $\mathcal{F}$ Mean for the DAVIS dataset and mean Intersection-over-Union (mIoU) for the FBMS benchmark. As Table~\ref{tab:segmentationComp} shows, `ours+CRF' denotes the proposed network with conditional random field (CRF)~\cite{krahenbuhl2011efficient} refinement which achieves the best $\mathcal{J}$ Mean and $\mathcal{F}$ Mean on DAVIS, and the best mIoU on FBMS. Our proposed method alone also demonstrate remarkable performance, the second best $\mathcal{F}$ Mean on DAVIS and the second best mIoU on FBMS.

\subsection{Effectiveness of the proposed network architecture}
In Table~\ref{tab:networkArchit}, we verify the effectiveness of the proposed dual-branch network architecture which deploys the proposed attention modules at both the encoder and the decoder side. `Appearance branch' denotes the appearance sub-network in Figure~\ref{fig:MultiTaskNetwork} while `motion branch' represents the motion sub-network. `Dual branch+MGA-D' is a model bridging two branches with the MGA module only at the decoder side, namely, the MGA-5. `Dual branch+MGA-E' consists of two branches with the MGA modules at the encoder side, namely, MGA-\{0-4\}. As Table~\ref{tab:networkArchit} indicates,
the dual branch+MGA-D outperforms the appearance branch by 6.6\% S-m and the motion branch by 14.4\% S-m on FBMS. The dual branch+MGA-E exceeds the appearance sub-network by 3.4\% maxF and the motion one by 8.6\% maxF on DAVIS.
The above statistics suggest that placing the attention modules at either the encoder or the decoder side can improve our proposed dual-branch architecture. The proposed network with attention modules at both the encoder and the decoder side surpasses the dual branch+MGA-D by 1.9\% maxF and the dual branch+MGA-E by 1.7\% maxF on FBMS. It implies that deploying MGA modules at encoder and decoder can slightly complements each other, and further enhances the performance.
\begin{table}[]
%\footnotesize
%\scriptsize
\tiny
\begin{center}
\resizebox{\linewidth}{!}{
    \begin{tabular}{|l|r|rr|r|}
        \hline
        \multirow{2}{*}{Methods}& \multirow{2}{*}{Year}& \multicolumn{2}{c|}{DAVIS} & FBMS     \\
                                &                      & $\mathcal{J}$ Mean   & $\mathcal{F}$ Mean & mIoU \\
        \hline
        SAGE~\cite{wang2015saliency}     & CVPR'15 & 41.5     & 36.9 & 61.2 \\
        LVO~\cite{tokmakov2017learning}  & ICCV'17 & 75.9     & 72.1 & 65.1 \\
        FSEG~\cite{jain2017fusionseg}    & CVPR'17 & 70.7     & 65.3 & 68.4 \\
        ARP~\cite{koh2017primary}        & CVPR'17 & 76.2     & 70.2 & 59.8 \\
        PDB~\cite{song2018pyramid}       & ECCV'18 & 74.3     & 72.8 & 72.3 \\
        PDB+CRF  & ECCV'18 & 77.2     & 74.5 & {\color{blue}74.0} \\
        MSGSTP~\cite{hu2018unsupervised} & ECCV'18 & 77.6     & 75.0 & 60.8 \\
        MBN~\cite{li2018unsupervised}    & ECCV'18 & {\color{green}80.4} & {\color{blue}78.5} & 73.9 \\
        %FGRNE    & CVPR'18 & 67.5     & 64.5 & 65.0 \\
        IET~\cite{li2018instance}        & CVPR'18 & 78.6     & 76.1 & 71.9 \\
        MotAdapt~\cite{siam2018video}    & ICRA'19 & 77.2     & 77.4 & \rule[1.0mm]{0.45cm}{0.1mm}     \\
        \hline
        ours     &         & {\color{blue}80.2} & {\color{green}80.8} & {\color{green}82.6} \\
        ours+CRF &         & {\color{red}81.4}  & {\color{red}81.0} & {\color{red}82.8} \\
        \hline
    \end{tabular}
}
\end{center}
\caption{Comparisons with state-of-the-art unsupervised video segmentation algorithms. The three
best performing algorithms are marked with {\color{red}red}, {\color{green}green} and {\color{blue}blue} colors respectively.}
\label{tab:segmentationComp}
\end{table}

\begin{table}[t]
\begin{center}
\resizebox{\linewidth}{!}{
    \begin{tabular}{|l | rrr | rrr|}
    \hline
    \multirow{2}{*}{Methods} &                & DAVIS          &                &       & FBMS  &       \\
                             & MAE            & S-m             & maxF           & MAE   & S-m    & maxF  \\
    \hline
    appearance branch        & 0.031          & 0.882          & 0.865          & 0.094          & 0.833 & 0.867 \\
    motion branch            & 0.035          & 0.859          & 0.813          & 0.083          & 0.755 & 0.767 \\
    dual branch+MGA-D        & 0.024          & 0.900          & 0.889          & 0.029          & 0.899 & 0.891 \\
    dual branch+MGA-E        & \textbf{0.021} & 0.913          & 0.899          & 0.030          & 0.903 & 0.893 \\
    ours                     & 0.022          & \textbf{0.913} & \textbf{0.902} & \textbf{0.027} & \textbf{0.907} & \textbf{0.910} \\
    \hline
    \end{tabular}
}
\end{center}
\caption{Effectiveness of the proposed network architecture.}
\label{tab:networkArchit}
\end{table}

\subsection{Effectiveness of the proposed motion guided attention}
To explore the effectiveness of our proposed motion guided attention modules, we compare the MGA modules with some naive fusions including concatenation, elementwise multiplication and addition which are denoted as `Concat', `Mul' and `Add' respectively in Table~\ref{tab:compNaiveFusion}. Specifically, The Concat fusion first concatenates a $C$-channel appearance feature and a $C'$-channel motion feature/map along the depth dimension, and then applies a $1\times 1$ convolution with $C$ output channels. To fuse two tensors, the Mul module first adjust a $C'$-channel motion feature to be $C$-channel via a $1\times 1$ convolution, and then elementwisely multiplies the motion feature with a $C$-channel appearance feature. To fuse a tensor and a map, the Mul module multiplies each channel slice of an appearance feature by a motion saliency map. The Add fusion works in a way similar to the Mul fusion. For the Concat, Mul and Add fusion in Table~\ref{tab:compNaiveFusion}, their corresponding fusion operator respectively replaces the MGA-\{0-5\} in Figure~\ref{fig:MultiTaskNetwork} to form their own model. As can be seen in Table~\ref{tab:compNaiveFusion}, our proposed motion guided attention modules surpasses the best naive fusion `Add' by 2.4\% S-m and 3.8\% maxF on DAVIS, which suggests that the proposed MGA modules effectively integrate the appearance and the motion branch.

\begin{table}[]
\footnotesize
\begin{center}
\scriptsize
\resizebox{\linewidth}{!}{
    \begin{tabular}{|l | rrr | rrr|}
    \hline
    \multirow{2}{*}{Methods} &       & DAVIS &       &       & FBMS  &       \\
                             & MAE   & S-m    & maxF  & MAE   & S-m    & maxF  \\
    \hline
    Concat                   & 0.030          & 0.876          & 0.844          & 0.068 & 0.815 & 0.822 \\
    Mul                      & 0.030          & 0.877          & 0.847          & 0.079 & 0.785 & 0.810 \\
    Add                      & 0.027          & 0.891          & 0.864          & 0.040 & 0.888 & 0.898 \\
    ours                     & \textbf{0.022} & \textbf{0.913} & \textbf{0.902} & \textbf{0.027} & \textbf{0.907} & \textbf{0.910} \\
    \hline
    \end{tabular}
}
\end{center}
\caption{Comparison with naive fusions.}
\label{tab:compNaiveFusion}
\end{table}

As shown in Table~\ref{tab:attentionEncoderDecoder}, we separately verify the effectiveness of the proposed MGA-m, MGA-t, MGA-tm and MGA-tmc. `E-$*$' denotes deploying the attention or fusion module $*$ at the encoder, specifically, the positions of MGA-\{0-4\}. `D-$*$' refers to placing the module $*$ at the decoder side, namely, the position of MGA-5. `*-Concat', `*-Mul' and `*-Add' are implemented as the same way as those in Table~\ref{tab:compNaiveFusion}. For `E-$*$' models, their attention module at the decoder side is chosen as MGA-m. For `D-$*$' models, their attention type at the encoder side is MGA-tmc. As Table~\ref{tab:compNaiveFusion} displays, all of our proposed MGA modules outperform the naive fusions. For examples, E-MGA-tm surpasses E-Add by 1.7\% maxF on DAVIS and E-MGA-t obtains 1.6\% S-m higher than E-Add on FBMS. At the encoder side, the MGA-tmc module achieve the best results. As for the decoder side, the MGA-m achieves the highest accuracy, which exceeds D-Add by 0.9\% S-m and 0.8\% maxF on FBMS.

\subsection{Effectiveness of the proposed training scheme}
\begin{table}[]
	\begin{center}
		\scriptsize
		\resizebox{\linewidth}{!}{
			\begin{tabular}{|l | rrr | rrr|}
				\hline
				\multirow{2}{*}{Methods} &       & DAVIS &       &       & FBMS  &       \\
				& MAE   & S-m    & maxF  & MAE   & S-m    & maxF  \\
				\hline
				E-Concat                 & 0.030 & 0.880 & 0.845 & 0.060 & 0.828 & 0.841 \\
				E-Mul                    & 0.032 & 0.873 & 0.846 & 0.082 & 0.786 & 0.804 \\
				E-Add                    & 0.026 & 0.895 & 0.876 & 0.038 & 0.890 & 0.893 \\
				E-MGA-t                  & 0.023 & 0.907 & 0.899 & 0.030 & 0.906 & 0.901 \\
				E-MGA-tm                 & 0.026 & 0.902 & 0.893 & 0.028 & 0.906 & 0.907 \\
				E-MGA-tmc                & \textbf{0.022} & \textbf{0.913} & \textbf{0.902} & \textbf{0.027} & \textbf{0.907} & \textbf{0.910} \\
				\hline
				D-Concat                 & 0.024          & 0.904          & 0.894 & 0.030 & 0.902 & 0.894 \\
				D-Mul                    & \textbf{0.021} & 0.913          & 0.900 & 0.029 & 0.904 & 0.900 \\
				D-Add                    & 0.023          & 0.907          & 0.899 & 0.033 & 0.898 & 0.902 \\
				D-MGA-m                  & 0.022          & \textbf{0.913} & \textbf{0.902} & \textbf{0.027} & \textbf{0.907} & \textbf{0.910} \\
				\hline
			\end{tabular}
		}
	\end{center}
	\caption{Effectiveness of the proposed motion guided attention modules at encoder and decoder side.}
	\label{tab:attentionEncoderDecoder}
\end{table}
We investigate whether it is beneficial to divide the video SOD task into two sub-tasks, and to solve these sub-tasks in advance. As shown in Table~\ref{tab:multiTaskTraining}, $T_0$ denotes a training scheme that do not train the appearance branch on static-image salient object detection or train the motion branch on motion saliency detection beforehand. The $T_0$ method initializes the encoders with pretrained image classification models~\cite{he2016deep}, randomizes other parameters, and trains the whole proposed network on the video SOD task. Different from $T_0$, the $T_m$ scheme pretrains the motion branch alone on the motion saliency detection sub-task, while the $T_a$ method pretrains the appearance branch on static-image SOD sub-task. $T_{ma}$ represents our proposed multi-task training scheme which separately tunes the two branches on their corresponding sub-task before end-to-end training the whole network. As Table~\ref{tab:multiTaskTraining} displays, $T_{ma}$ exceeds the second best $T_a$ by 1.7\% maxF on DAVIS and 1.9\% maxF on FBMS, which suggests that our proposed multi-task training scheme helps capture more accurate features. Note that $T_m$ demonstrates better results on DAVIS but worse performance on FBMS, in comparison to $T_0$. It may be due to that the videos from FBMS usually contains multiple salient objects and not all these objects have discriminative motion pattern. Thus the $T_m$ model, which only has been pretrained to locate salient motions, could be over-reliant on the motion cues to some degree, and struggles to harvest more accurate appearance contrast.

\begin{table}[]
\begin{center}
\resizebox{\linewidth}{!}{
    \begin{tabular}{|l|cc|rrr|rrr|}
    \hline
    \multirow{2}{*}{Methods} & \multirow{2}{*}{\begin{tabular}[c]{@{}l@{}}pretrain\\ appearance ?\end{tabular}} & \multirow{2}{*}{\begin{tabular}[c]{@{}l@{}}pretrain\\ motion ?\end{tabular}} &       & DAVIS &       &       & FBMS  &       \\
                             &                                                                              &                                                                              & MAE   & S-m    & maxF  & MAE   & S-m    & maxF  \\
    \hline
    $T_0$ & $\times$   & $\times$   & 0.043 & 0.870 & 0.859 & 0.036 & 0.893 & 0.879 \\
    $T_m$ & $\times$   & \checkmark & 0.026 & 0.892 & 0.873 & 0.059 & 0.835 & 0.856 \\
    $T_a$ & \checkmark & $\times$   & 0.025 & 0.897 & 0.885 & 0.035 & 0.896 & 0.881 \\
    $T_{ma}$ & \checkmark& \checkmark & \textbf{0.022} & \textbf{0.913} & \textbf{0.902} & \textbf{0.027} & \textbf{0.907} & \textbf{0.910} \\
    \hline
    \end{tabular}
}
\end{center}
\caption{Effectiveness of the proposed multi-task training scheme.}
\label{tab:multiTaskTraining}
\end{table}

%------------------------------------------------------------------------
\section{Conclusions}
This paper introduces a novel motion guided attention network which sets up a new state-of-the-art baseline for the video salient object detection task. To the best of our knowledge, the proposed network is the first to successfully model how salient motion patterns affect object saliency in an attention scheme. The proposed motion guided attention modules effectively instantiate such attention mechanism to model the influence from salient motions to visual saliency. Using motion cues resulting from the previous frame, our proposed method sufficiently exploits temporal context, superior to existing long-range memory based models.

\section*{Acknowledgment}
This work was supported by the Hong Kong PhD Fellowship, the National Natural Science Foundation of China under Grant No.U1811463 and No.61702565, the Fundamental Research Funds for the Central Universities under Grant No.18lgpy63, and was also sponsored by SenseTime Research Fund.

{\small
\bibliographystyle{ieee_fullname}
\bibliography{egbib}

\begin{thebibliography}{10}\itemsep=-1pt

\bibitem{achanta2009frequency}
Radhakrishna Achanta, Sheila Hemami, Francisco Estrada, and Sabine Susstrunk.
\newblock Frequency-tuned salient region detection.
\newblock In {\em Proceedings of the IEEE Conference on Computer Vision and
  Pattern Recognition (CVPR)}, pages 1597--1604, 2009.

\bibitem{brox2010object}
Thomas Brox and Jitendra Malik.
\newblock Object segmentation by long term analysis of point trajectories.
\newblock In {\em Proceedings of the European Conference on Computer Vision
  (ECCV)}, pages 282--295. Springer, 2010.

\bibitem{Butler:ECCV:2012}
D.~J. Butler, J. Wulff, G.~B. Stanley, and M.~J. Black.
\newblock A naturalistic open source movie for optical flow evaluation.
\newblock In {A. Fitzgibbon et al. (Eds.)}, editor, {\em Proceedings of the
  European Conference on Computer Vision (ECCV)}, Part IV, LNCS 7577, pages
  611--625. Springer-Verlag, Oct. 2012.

\bibitem{chen2017video}
Chenglizhao Chen, Shuai Li, Yongguang Wang, Hong Qin, and Aimin Hao.
\newblock Video saliency detection via spatial-temporal fusion and low-rank
  coherency diffusion.
\newblock {\em IEEE Transactions on Image Processing}, 26(7):3156--3170, 2017.

\bibitem{chen2018reverse}
Shuhan Chen, Xiuli Tan, Ben Wang, and Xuelong Hu.
\newblock Reverse attention for salient object detection.
\newblock In {\em Proceedings of the European Conference on Computer Vision
  (ECCV)}, pages 234--250, 2018.

\bibitem{deng2009imagenet}
Jia Deng, Wei Dong, Richard Socher, Li-Jia Li, Kai Li, and Li Fei-Fei.
\newblock Imagenet: A large-scale hierarchical image database.
\newblock In {\em IEEE Conference on Computer Vision and Pattern Recognition
  (CVPR)}, pages 248--255, 2009.

\bibitem{deng2018r3net}
Zijun Deng, Xiaowei Hu, Lei Zhu, Xuemiao Xu, Jing Qin, Guoqiang Han, and
  Pheng-Ann Heng.
\newblock R3net: Recurrent residual refinement network for saliency detection.
\newblock In {\em Proceedings of the 27th International Joint Conference on
  Artificial Intelligence (IJCAI)}, pages 684--690. AAAI Press, 2018.

\bibitem{dosovitskiy2015flownet}
Alexey Dosovitskiy, Philipp Fischery, Eddy Ilg, Philip Hausser, Caner Hazirbas,
  V Golkov, Patrick~Van Der~Smagt, Daniel Cremers, and Thomas Brox.
\newblock Flownet: Learning optical flow with convolutional networks.
\newblock In {\em Proceedings of the IEEE International Conference on Computer
  Vision (ICCV)}, pages 2758--2766, 2015.

\bibitem{fan2017structure}
Deng-Ping Fan, Ming-Ming Cheng, Yun Liu, Tao Li, and Ali Borji.
\newblock Structure-measure: A new way to evaluate foreground maps.
\newblock In {\em Proceedings of the IEEE International Conference on Computer
  Vision (ICCV)}, pages 4558--4567, 2017.

\bibitem{Fan_2019_CVPR}
Deng-Ping Fan, Wenguan Wang, Ming-Ming Cheng, and Jianbing Shen.
\newblock Shifting more attention to video salient object detection.
\newblock In {\em The IEEE Conference on Computer Vision and Pattern
  Recognition (CVPR)}, June 2019.

\bibitem{fu2017look}
Jianlong Fu, Heliang Zheng, and Tao Mei.
\newblock Look closer to see better: Recurrent attention convolutional neural
  network for fine-grained image recognition.
\newblock In {\em Proceedings of the IEEE Conference on Computer Vision and
  Pattern Recognition (CVPR)}, volume~2, page~3, 2017.

\bibitem{he2016deep}
Kaiming He, Xiangyu Zhang, Shaoqing Ren, and Jian Sun.
\newblock Deep residual learning for image recognition.
\newblock In {\em Proceedings of the IEEE Conference on Computer Vision and
  Pattern Recognition (CVPR)}, pages 770--778, 2016.

\bibitem{hou2017deeply}
Qibin Hou, Ming-Ming Cheng, Xiaowei Hu, Ali Borji, Zhuowen Tu, and Philip~HS
  Torr.
\newblock Deeply supervised salient object detection with short connections.
\newblock In {\em Proceedings of the IEEE Conference on Computer Vision and
  Pattern Recognition (CVPR)}, pages 3203--3212, 2017.

\bibitem{hu2018unsupervised}
Yuan-Ting Hu, Jia-Bin Huang, and Alexander~G Schwing.
\newblock Unsupervised video object segmentation using motion saliency-guided
  spatio-temporal propagation.
\newblock In {\em Proceedings of the European Conference on Computer Vision
  (ECCV)}, pages 786--802, 2018.

\bibitem{ilg2017flownet}
Eddy Ilg, Nikolaus Mayer, Tonmoy Saikia, Margret Keuper, Alexey Dosovitskiy,
  and Thomas Brox.
\newblock Flownet 2.0: Evolution of optical flow estimation with deep networks.
\newblock In {\em Proceedings of the IEEE Conference on Computer Vision and
  Pattern Recognition (CVPR)}, pages 1647--1655, 2017.

\bibitem{itti2004automatic}
Laurent Itti.
\newblock Automatic foveation for video compression using a neurobiological
  model of visual attention.
\newblock {\em IEEE Transactions on Image Processing}, 13(10):1304--1318, 2004.

\bibitem{itti1998a}
Laurent Itti, Christof Koch, and Ernst Niebur.
\newblock A model of saliency-based visual attention for rapid scene analysis.
\newblock {\em IEEE Transactions on Pattern Analysis and Machine Intelligence},
  20(11):1254--1259, 1998.

\bibitem{jain2017fusionseg}
Suyog~Dutt Jain, Bo Xiong, and Kristen Grauman.
\newblock Fusionseg: Learning to combine motion and appearance for fully
  automatic segmentation of generic objects in videos.
\newblock In {\em Proceedings of the IEEE Conference on Computer Vision and
  Pattern Recognition (CVPR)}, pages 2117--2126, 2017.

\bibitem{koh2017primary}
Yeong~Jun Koh and Chang-Su Kim.
\newblock Primary object segmentation in videos based on region augmentation
  and reduction.
\newblock In {\em 2017 IEEE Conference on Computer Vision and Pattern
  Recognition (CVPR)}, pages 7417--7425. IEEE, 2017.

\bibitem{krahenbuhl2011efficient}
Philipp Kr{\"a}henb{\"u}hl and Vladlen Koltun.
\newblock Efficient inference in fully connected crfs with gaussian edge
  potentials.
\newblock In {\em Advances in Neural Information Processing Systems}, pages
  109--117, 2011.

\bibitem{le2018video}
Trung-Nghia Le and Akihiro Sugimoto.
\newblock Video salient object detection using spatiotemporal deep features.
\newblock {\em IEEE Transactions on Image Processing}, 27(10):5002--5015, 2018.

\bibitem{li2017instance}
Guanbin Li, Yuan Xie, Liang Lin, and Yizhou Yu.
\newblock Instance-level salient object segmentation.
\newblock In {\em Proceedings of the IEEE Conference on Computer Vision and
  Pattern Recognition (CVPR)}, pages 2386--2395, 2017.

\bibitem{li2018flow}
Guanbin Li, Yuan Xie, Tianhao Wei, Keze Wang, and Liang Lin.
\newblock Flow guided recurrent neural encoder for video salient object
  detection.
\newblock In {\em Proceedings of the IEEE Conference on Computer Vision and
  Pattern Recognition (CVPR)}, pages 3243--3252, 2018.

\bibitem{li2018instance}
Siyang Li, Bryan Seybold, Alexey Vorobyov, Alireza Fathi, Qin Huang, and C-C
  Jay~Kuo.
\newblock Instance embedding transfer to unsupervised video object
  segmentation.
\newblock In {\em Proceedings of the IEEE Conference on Computer Vision and
  Pattern Recognition (CVPR)}, pages 6526--6535, 2018.

\bibitem{li2018unsupervised}
Siyang Li, Bryan Seybold, Alexey Vorobyov, Xuejing Lei, and C-C Jay~Kuo.
\newblock Unsupervised video object segmentation with motion-based bilateral
  networks.
\newblock In {\em Proceedings of the European Conference on Computer Vision
  (ECCV)}, pages 207--223, 2018.

\bibitem{li2018contour}
Xin Li, Fan Yang, Hong Cheng, Wei Liu, and Dinggang Shen.
\newblock Contour knowledge transfer for salient object detection.
\newblock In {\em Proceedings of the European Conference on Computer Vision
  (ECCV)}, pages 355--370, 2018.

\bibitem{liu2018picanet}
Nian Liu, Junwei Han, and Ming-Hsuan Yang.
\newblock Picanet: Learning pixel-wise contextual attention for saliency
  detection.
\newblock In {\em Proceedings of the IEEE Conference on Computer Vision and
  Pattern Recognition (CVPR)}, pages 3089--3098, 2018.

\bibitem{liu2017saliency}
Zhi Liu, Junhao Li, Linwei Ye, Guangling Sun, and Liquan Shen.
\newblock Saliency detection for unconstrained videos using superpixel-level
  graph and spatiotemporal propagation.
\newblock {\em IEEE Transactions on Circuits and Systems for Video Technology},
  27(12):2527--2542, 2017.

\bibitem{luo2017non}
Zhiming Luo, Akshaya Mishra, Andrew Achkar, Justin Eichel, Shaozi Li, and
  Pierre-Marc Jodoin.
\newblock Non-local deep features for salient object detection.
\newblock In {\em Proceedings of the IEEE Conference on Computer Vision and
  Pattern Recognition (CVPR)}, pages 6609--6617, 2017.

\bibitem{perazzi2016benchmark}
Federico Perazzi, Jordi Pont-Tuset, Brian McWilliams, Luc Van~Gool, Markus
  Gross, and Alexander Sorkine-Hornung.
\newblock A benchmark dataset and evaluation methodology for video object
  segmentation.
\newblock In {\em Proceedings of the IEEE Conference on Computer Vision and
  Pattern Recognition (CVPR)}, pages 724--732, 2016.

\bibitem{russakovsky2015imagenet}
Olga Russakovsky, Jia Deng, Hao Su, Jonathan Krause, Sanjeev Satheesh, Sean Ma,
  Zhiheng Huang, Andrej Karpathy, Aditya Khosla, Michael Bernstein, et~al.
\newblock Imagenet large scale visual recognition challenge.
\newblock {\em International Journal of Computer Vision}, 115(3):211--252,
  2015.

\bibitem{siam2018video}
Mennatullah Siam, Chen Jiang, Steven Lu, Laura Petrich, Mahmoud Gamal, Mohamed
  Elhoseiny, and Martin Jagersand.
\newblock Video segmentation using teacher-student adaptation in a human robot
  interaction (hri) setting.
\newblock In {\em IEEE International Conference on Robotics and Automation
  (ICRA)}, 2019.

\bibitem{song2018pyramid}
Hongmei Song, Wenguan Wang, Sanyuan Zhao, Jianbing Shen, and Kin-Man Lam.
\newblock Pyramid dilated deeper convlstm for video salient object detection.
\newblock In {\em Proceedings of the European Conference on Computer Vision
  (ECCV)}, pages 715--731, 2018.

\bibitem{sun2018pwc-net}
Deqing Sun, Xiaodong Yang, Mingyu Liu, and Jan Kautz.
\newblock Pwc-net: Cnns for optical flow using pyramid, warping, and cost
  volume.
\newblock In {\em Proceedings of the IEEE Conference on Computer Vision and
  Pattern Recognition (CVPR)}, pages 8934--8943, 2018.

\bibitem{sun2018sg}
Meijun Sun, Ziqi Zhou, Qinghua Hu, Zheng Wang, and Jianmin Jiang.
\newblock Sg-fcn: A motion and memory-based deep learning model for video
  saliency detection.
\newblock {\em IEEE Transactions on Cybernetics}, (99):1--12, 2018.

\bibitem{tokmakov2017learningmotion}
Pavel Tokmakov, Karteek Alahari, and Cordelia Schmid.
\newblock Learning motion patterns in videos.
\newblock In {\em Proceedings of the IEEE Conference on Computer Vision and
  Pattern Recognition (CVPR)}, pages 3386--3394, 2017.

\bibitem{tokmakov2017learning}
Pavel Tokmakov, Karteek Alahari, and Cordelia Schmid.
\newblock Learning video object segmentation with visual memory.
\newblock In {\em Proceedings of the IEEE International Conference on Computer
  Vision (ICCV)}, pages 4481--4490, 2017.

\bibitem{wang2017residual}
Fei Wang, Mengqing Jiang, Chen Qian, Shuo Yang, Cheng Li, Honggang Zhang,
  Xiaogang Wang, and Xiaoou Tang.
\newblock Residual attention network for image classification.
\newblock In {\em Proceedings of the IEEE Conference on Computer Vision and
  Pattern Recognition (CVPR)}, pages 6450--6458, 2017.

\bibitem{wang2017learning}
Lijun Wang, Huchuan Lu, Yifan Wang, Mengyang Feng, Dong Wang, Baocai Yin, and
  Xiang Ruan.
\newblock Learning to detect salient objects with image-level supervision.
\newblock In {\em Proceedings of the IEEE Conference on Computer Vision and
  Pattern Recognition (CVPR)}, pages 136--145, 2017.

\bibitem{wang2017stagewise}
Tiantian Wang, Ali Borji, Lihe Zhang, Pingping Zhang, and Huchuan Lu.
\newblock A stagewise refinement model for detecting salient objects in images.
\newblock In {\em Proceedings of the IEEE International Conference on Computer
  Vision (ICCV)}, pages 4019--4028, 2017.

\bibitem{wang2018detect}
Tiantian Wang, Lihe Zhang, Shuo Wang, Huchuan Lu, Gang Yang, Xiang Ruan, and
  Ali Borji.
\newblock Detect globally, refine locally: A novel approach to saliency
  detection.
\newblock In {\em Proceedings of the IEEE Conference on Computer Vision and
  Pattern Recognition (CVPR)}, pages 3127--3135, 2018.

\bibitem{wang2015saliency}
Wenguan Wang, Jianbing Shen, and Fatih Porikli.
\newblock Saliency-aware geodesic video object segmentation.
\newblock In {\em Proceedings of the IEEE Conference on Computer Vision and
  Pattern Recognition (CVPR)}, pages 3395--3402, 2015.

\bibitem{wang2015consistent}
Wenguan Wang, Jianbing Shen, and Ling Shao.
\newblock Consistent video saliency using local gradient flow optimization and
  global refinement.
\newblock {\em IEEE Transactions on Image Processing}, 24(11):4185--4196, 2015.

\bibitem{wang2018video}
Wenguan Wang, Jianbing Shen, and Ling Shao.
\newblock Video salient object detection via fully convolutional networks.
\newblock {\em IEEE Transactions on Image Processing}, 27(1):38--49, 2018.

\bibitem{wu2014weighted}
Hefeng Wu, Guanbin Li, and Xiaonan Luo.
\newblock Weighted attentional blocks for probabilistic object tracking.
\newblock {\em The Visual Computer}, 30(2):229--243, 2014.

\bibitem{wu2018interpretable}
Xian Wu, Guanbin Li, Qingxing Cao, Qingge Ji, and Liang Lin.
\newblock Interpretable video captioning via trajectory structured
  localization.
\newblock In {\em Proceedings of the IEEE Conference on Computer Vision and
  Pattern Recognition (CVPR)}, pages 6829--6837, 2018.

\bibitem{xu2015show}
Kelvin Xu, Jimmy Ba, Ryan Kiros, Kyunghyun Cho, Aaron Courville, Ruslan
  Salakhudinov, Rich Zemel, and Yoshua Bengio.
\newblock Show, attend and tell: Neural image caption generation with visual
  attention.
\newblock In {\em International Conference on Machine Learning (ICML)}, pages
  2048--2057, 2015.

\bibitem{zhang2017amulet}
Pingping Zhang, Dong Wang, Huchuan Lu, Hongyu Wang, and Xiang Ruan.
\newblock Amulet: Aggregating multi-level convolutional features for salient
  object detection.
\newblock In {\em Proceedings of the IEEE International Conference on Computer
  Vision (ICCV)}, pages 202--211, 2017.

\bibitem{zhang2017learning}
Pingping Zhang, Dong Wang, Huchuan Lu, Hongyu Wang, and Baocai Yin.
\newblock Learning uncertain convolutional features for accurate saliency
  detection.
\newblock In {\em Proceedings of the IEEE International Conference on Computer
  Vision (ICCV)}, pages 212--221, 2017.

\bibitem{zhao2013unsupervised}
Rui Zhao, Wanli Ouyang, and Xiaogang Wang.
\newblock Unsupervised salience learning for person re-identification.
\newblock In {\em Proceedings of the IEEE Conference on Computer Vision and
  Pattern Recognition (CVPR)}, pages 3586--3593, 2013.

\end{thebibliography}
}

\end{document}